\documentclass{article}

\usepackage{arxiv}
\usepackage{graphicx}
\usepackage{xcolor}
\usepackage{amsmath}
\usepackage{caption}
\usepackage{subcaption}
\usepackage[utf8]{inputenc} % allow utf-8 input
\usepackage[T1]{fontenc}    % use 8-bit T1 fonts
\usepackage{hyperref}       % hyperlinks
\usepackage{url}            % simple URL typesetting
\usepackage{booktabs}       % professional-quality tables
\usepackage{amsfonts}       % blackboard math symbols
\usepackage{nicefrac}       % compact symbols for 1/2, etc.
\usepackage{microtype}      % microtypography
\usepackage{lipsum}

\title{L-CNN: A Lattice cross-fusion strategy for multistream convolutional neural networks}

\author{Ana Paula G. S. de Almeida and Fl{\'a}vio de Barros Vidal}

\begin{document}

\maketitle

\abstract{This paper proposes a fusion strategy for multistream convolutional networks, the Lattice Cross Fusion. This approach crosses signals from convolution layers performing mathematical operation-based fusions right before pooling layers. Results on a purposely worsened CIFAR-10, a popular image classification data set, with a modified AlexNet-LCNN version show that this novel method outperforms by $46\%$ the baseline single stream network, with faster convergence, stability, and robustness.}

\section{Introduction}

Multistream Convolutional Neural Networks (also known as Multichannel Convolutional Neural Networks) - MCNN - has been developed, applied, and used in many situations and applications nowadays~\cite{1, 2, 3, 4, 5, 6, 7, 8, 9}. This model architecture is derived from traditional Convolutional Neural Networks (CNNs) proposed by \cite{10} and allow to employ basically all, traditional (or not) models available in the literature, as \textit{LeNet}\cite{10}, \textit{Alexnet}\cite{11}, \textit{VGG}\cite{12} and many others models, basically adjusting (or modifying) the fusion stage\cite{1}.

Usually, works that address application issues to a M-CNN approach do not focus on the fusion stage (e.g.~\cite{2, 12}). However, there are some efforts~\cite{3,13,14,15} to increase the performance of the networks. Gamulle \textit{et al.}~\cite{3} and Tu \textit{et al.}~\cite{13} uses a multi-stream approach to recognize human actions from video sequences. The former focus on learning salient spatial features and mapping their temporal relationship with the aid of Long-Short-Term-Memory (LSTM) networks and uses two different fusion methods: averaging and training a multi-class linear SVM using the softmax scores as features. The latter construct an appearance and a motion stream, concatenating the streams in a fusion module based on spatio-temporal 3D convolutions. Azar \textit{et al.}~\cite{14} use M-CNNs to recognize group activities and use concatenations to fuse all its streams. In~\cite{15}, M-CNNs are used to classificate high-resolution aerial scenes and two fusion techniques are evaluated, concatenation, and addition.

Karpathy \textit{et al.}~\cite{1} addressed the fusion issue to a novel model, placing two separate single-frame networks time-delayed apart and merging their outputs in a fusion step allowing improvements in the accuracy scores on video action recognition tasks. As described in Feichtenhofer \textit{et al.} ~\cite{4} the fusion stage can be performed at a convolutional layer without loss of performance (accuracy). Based on this finding, we propose in this manuscript a novel model based on a new cross-fusion method applied for all available models of MCNNs. This new proposed cross-fusion method is focused on observed features, specifically in the \textit{ReLu's} stage, used by many traditional MCNNs models developed in all known approaches (e.g. in \cite{1,16,4}).

\section{Proposed Model}
Our novel proposed model is based on a crossing signal inference among each data streaming (or channel) output from the convolution stage and processed by the ReLU's stage, connecting each of this output with others outputs coming from all others data streamings (or channels) using a new fusion function approaching. This model is formally described in the following Subsection. 

\subsection{Lattice Cross-fusion Strategy}
\begin{figure}
  \centering
  \includegraphics[width=\textwidth]{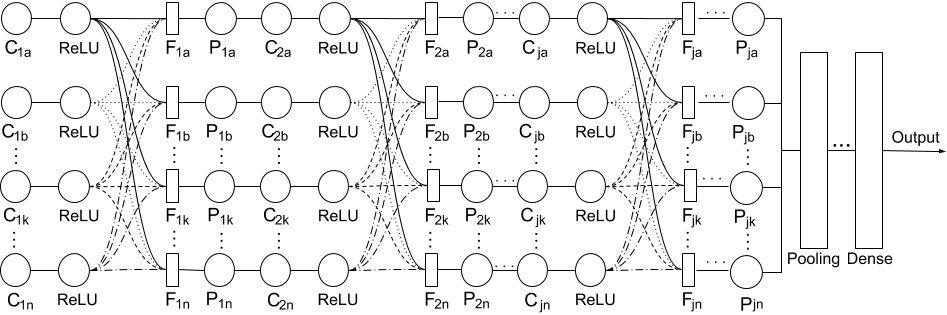}
  \caption{A general LCNN model. Convolutional layers are represented by $C$, the fusion modules are $F$ and the pooling layers are expressed by $P$.}
  \label{fig_1}
\end{figure}

The cross-fusion function $f: C_{a}, C_{b},..., C_{k},... C_{n} \rightarrow y$ combines the $n-1$ convolutional layers with the $C_{k} \in \mathbb{R}^{h \times w \times d}$  where  $h, w$ and $d$ are the height, width, and depth (number of channels/streams), respectively. The cross-fusion average fusion is defined as $y^{\textrm{avg}} = f^{\textrm{avg}}(C_{a}, C_{b}, ..., C_{k},..., C_{n})$. It computes the average of the $n$ convolutional layers, connecting as an input of the next pooling layer. It is important to point out that other types of mathematical operations can be applied in convolutional layers, such as addition, subtraction, absolute difference. These fusion stage inclusion processes are repeated along all CNNs' \textit{convolution-ReLU-dropout} layers sets, finishing before the dense stage. A visual description of this proposed cross-function strategy is presented in Figure~\ref{fig_1}.

%\begin{figure}[!htpb]
%\centering{\includegraphics[width=.5\textwidth]{general_model_2.jpg}}
%\caption{A general LCNN model. Convolutional layers are represented by %$C$, the fusion modules are $F$ and the pooling layers are expressed by $P$.}
%\label{fig_1}
%\end{figure}

\section{Experimental Evaluation}

For the purpose of evaluating our approach, our cross-fusion function is set to an \textit{average} operation and the used baseline (single and dual stream) architecture is AlexNet~\cite{17}.

Our new AlexNet-LCNN is defined as \textit{C(96,11,4) $\rightarrow$ LF(average) $\rightarrow$ P(2) $\rightarrow$ C(256,11,1) $\rightarrow$ LF(average) $\rightarrow$ P(2) $\rightarrow$ C(384,3,1) $\rightarrow$ C(384,3,1) $\rightarrow$ C(256,3,1) $\rightarrow$ LF(average) $\rightarrow$ P(2) $\rightarrow$ FL $\rightarrow$ FC(4096) $\rightarrow$ D(0.4) $\rightarrow$ FC(4096) $\rightarrow$ D(0.4) $\rightarrow$ FC(10)}, such that \textit{C(d, f, s)} indicates a convolution layer with $d$ filters of spatial size $f \times f$, applied to the input with stride $s$. \textit{LF($\alpha$)} means the lattice fusion realized with an operation cross-fusion function. \textit{P(s)} is the pooling layer with stride $s$. \textit{FL} is a layer that flattens the input. \textit{FC(n)} is a fully connected layer with $n$ nodes. \textit{D(p)} is a dropout layer of $p$ as a dropout rate, used exclusively during the training step.

In order to compare performances, we also implemented a late fusion multistream AlexNet (AlexNet-MCNN), consisting of two independent streams that merge right before the first fully connected layer as highlighted in~\cite{1}.

Additionally, for comparison proposes, we also evaluated a cross-modal CNN (AlexNet-XCNN)~\cite{16}, which is a typically image-based approach that each of the input images receives its own CNN superlayer, with cross-connections inserted after the pooling operation, and full weight sharing in the fully connected layers. This model was developed to explore the crossing-signal inference and its accuracy performance will be used to compare with our novel proposed L-CNN model.

Results also considered the traditional single stream AlexNet architecture~\cite{17} for both of our inputs, described below.

\begin{figure}
    \centering
    \begin{subfigure}[t]{.35\textwidth}
      \centering
      \includegraphics[width=\linewidth]{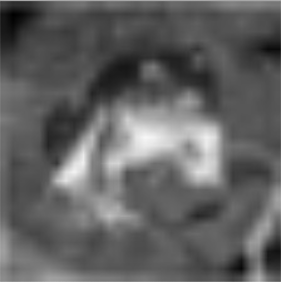}
      \caption{Grayscale image.}
      \label{fig_2:sub1}
    \end{subfigure}%
    ~
    \begin{subfigure}[t]{.35\textwidth}
      \centering
      \includegraphics[width=\linewidth]{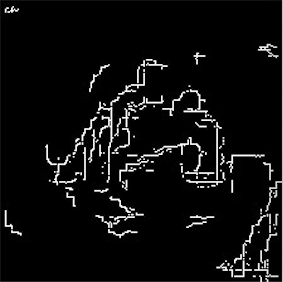}
      \caption{Image with edge detection.}
      \label{fig_2:sub2}
    \end{subfigure}
    \caption{A class sample from the CIFAR-10 data set~\cite{11}. In (a) the first stream, the input is a $224 \times 224$ grayscale image. In the second stream (b), the same image with edge detection.}
    \label{fig_2}
\end{figure}

The chosen test bed data set to evaluate our model was the CIFAR-10~\cite{11}, a popular image classification benchmark data set. For a multistreaming scenario, the chosen inputs were purposely worsened in order to evaluate the robustness increased by the proposed L-CNN model -- the main goal is not to achieve or improve state-of-the-art results. In the first stream, a grayscaled version of CIFAR-10 is used and in the second stream, an edge extraction -- created with Canny edge detector~\cite{18} -- of the first stream, presented respectively in Figures \ref{fig_2:sub1} and \ref{fig_2:sub2}. To respect AlexNet's original first convolutional stage constraints, both input streams were resized to $224 \times 224$.

Table~\ref{Table1} presents all the achieved accuracies with the previously described inputs and model comparisons when trained for 260 epochs. This number of epochs was chosen accordingly to loss and accuracy graphs presented in~\cite{16}. Given that the edge detection single stream presents constantly poor results in accuracy and loss, we can describe this input as a distractor to the network. Even the grayscale input did not perform well in a single stream scenario, as described in Table~\ref{Table1}. It also can be noticed that the L-CNN method outperforms all the other approaches, including the single stream models, using two low quality inputs and fewer features to learn, considering that we took the color information that would support the network in feature learning, in according to~\cite{19}.

\begin{table}[!htpb]
\centering
\caption{CIFAR-10 data set accuracies using a 2-streaming schema.}
{\begin{tabular}{|l|l|l|}\hline
Model & Loss & Accuracy\\\hline
AlexNet (grayscale) & 3.127 & 0.1664\\\hline
AlexNet (edge detection) & 12.9167 & 0.1008\\\hline
AlexNet-MCNN & 2.923 & 0.4853\\\hline
AlexNet-XCNN & 2.35 & 0.4896\\\hline
\textbf{AlexNet}-LCNN & \textbf{2.922} & \textbf{0.6257}\\\hline
\end{tabular}
\label{Table1}}{}
\end{table}

\begin{figure}
\centering
    \begin{subfigure}[t]{.5\textwidth}
      \centering
      \includegraphics[width=\linewidth]{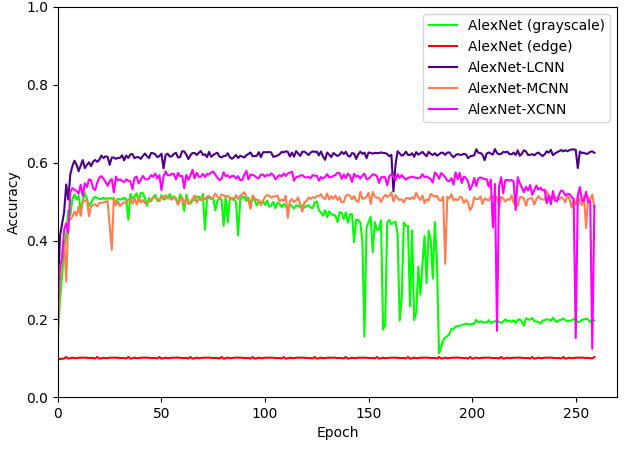}
      \caption{CIFAR-10 test accuracies.}
      \label{fig_3:sub1}
    \end{subfigure}
    \begin{subfigure}[t]{.5\textwidth}
      \centering
      \includegraphics[width=\linewidth]{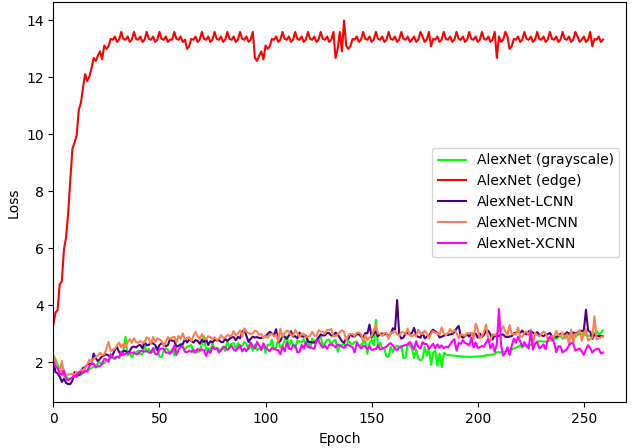}
      \caption{CIFAR-10 test losses.}
      \label{fig_3:sub2}
    \end{subfigure}
 \caption{CIFAR-10 test accuracies and losses under training epochs.}
 \label{fig_3}
 \end{figure}

The graphs shown in Figures ~\ref{fig_3:sub1} and ~\ref{fig_3:sub2} clearly confirms that our edge detection signal does not add consistent information to the network, as expected. Furthermore, the grayscale stream does not have a good performance by itself. Using this information we can also point out stability from the AlexNet-LCNN, achieving its peak about $50$ epochs and maintaining accuracy and loss during all the training process, unlike AlexNet-XCNN and even AlexNet(grayscale). Moreover, LF was able to increase the network robustness results by using poor signals as inputs.

\section{Conclusions and future works}
In this work, a Lattice Fusion is proposed for multistream convolutional neural networks. The proposed fusion uses mathematical operations before every pooling layer of a CNN architecture. Using an AlexNet network as a baseline and CIFAR-10 data set, we implemented three different model versions: a multistream CNN with late fusion, a cross-CNN, and our lattice-CNN, alongside two single streams traditional AlexNet. Experimental results show that the proposed LF outperformed all the aforementioned models at least $28\%$ when compared to all tested models. Also, the proposed fusion demonstrated robustness and stability, even when distractors are used as inputs. Future work will focus on working with different operations and data sets, examining how color spaces could improve our model accuracies, including new test bed scenarios, data sets, models and architectures of CNNs, and other multistream approaches available (e.g. LSTM, RNNs,...). Furthermore, additional streams will be analyzed and tested.
\vskip3pt

\vskip5pt

\noindent Ana Paula G. S. de Almeida and Fl{\'a}vio de Barros Vidal (\textit{University of Bras{\'i}lia, Bras{\'i}lia, Brazil})
\vskip3pt

\noindent E-mail: anapaula.gsa@gmail.com

\noindent Published Version available at \url{https://ieeexplore.ieee.org/document/8930458}. To cite this published version, please use the information as described below:

A. P. G. S. de Almeida and F. de Barros Vidal, "L-CNN: a lattice cross-fusion strategy for multistream convolutional neural networks," in Electronics Letters, vol. 55, no. 22, pp. 1180-1182, 31 10 2019, doi: 10.1049/el.2019.2631.

\bibliographystyle{unsrt}  
%\bibliography{references}

\end{document}